\documentclass[lettersize,journal]{IEEEtran}
\usepackage{amsmath,amsfonts}
\usepackage{algorithmic}
\usepackage{algorithm}
\usepackage{array}
\usepackage[caption=false,font=normalsize,labelfont=sf,textfont=sf]{subfig}
\usepackage{textcomp}
\usepackage{stfloats}
\usepackage{url}
\usepackage{verbatim}
\usepackage{tabularx}
\usepackage{graphicx}
\usepackage[switch]{lineno}
\usepackage{multicol}
\usepackage{multirow}
\usepackage{cite}
\usepackage[justification=centering]{caption}

\begin{document}

\title{EmoLat: Text-driven Image Sentiment Transfer via Emotion Latent Space}
\author{
    Jing Zhang\IEEEauthorrefmark{1},
    Bingjie Fan\IEEEauthorrefmark{1},
    Jixiang Zhu\IEEEauthorrefmark{1},
    Zhe Wang\IEEEauthorrefmark{1}
    \thanks{\IEEEauthorrefmark{1} All authors are with the department of Computer Science and Engineering, East China University of Science and Technology, Shanghai 200237, P. R. China. 
    \\ 
    E-mails: jingzhang@ecust.edu.cn, wangzhe@ecust.edu.cn}

}





\maketitle

\begin{abstract}
We propose EmoLat, a novel emotion latent space that enables fine-grained, text-driven image sentiment transfer by modeling cross-modal correlations between textual semantics and visual emotion features. Within EmoLat, an emotion semantic graph is constructed to capture the relational structure among emotions, objects, and visual attributes. To enhance the discriminability and transferability of emotion representations, we employ adversarial regularization, aligning the latent emotion distributions across modalities. Building upon EmoLat, a cross-modal sentiment transfer framework is proposed to manipulate image sentiment via joint embedding of text and EmoLat features. The network is optimized using a multi-objective loss incorporating semantic consistency, emotion alignment, and adversarial regularization. To support effective modeling, we construct EmoSpace Set, a large-scale benchmark dataset comprising images with dense annotations on emotions, object semantics, and visual attributes. Extensive experiments on EmoSpace Set demonstrate that our approach significantly outperforms existing state-of-the-art methods in both quantitative metrics and qualitative transfer fidelity, establishing a new paradigm for controllable image sentiment editing guided by textual input. The EmoSpace Set and all the code are available at http://github.com/JingVIPLab/EmoLat.
\end{abstract}


\section{Introduction}
\IEEEPARstart{I}{mage} sentiment transfer aims to transfer a target sentiment to a content image while preserving its original content, and ensure that the image exhibits visual effects consistent with the desired sentiment. Existing image sentiment transfer methods rely primarily on a target image to guide the transfer of emotion features. However, in real-world applications, a suitable target image may be difficult to obtain, significantly limiting the applicability of these kinds of methods. Compared to reference images, sentiment word can provide greater flexibility and accessibility, hence text-driven image sentiment transfer that uses a sentiment word to guide the emotion transfer receives more attention. 

\begin{figure}[!t]
  \centering
  \includegraphics[width=\linewidth]{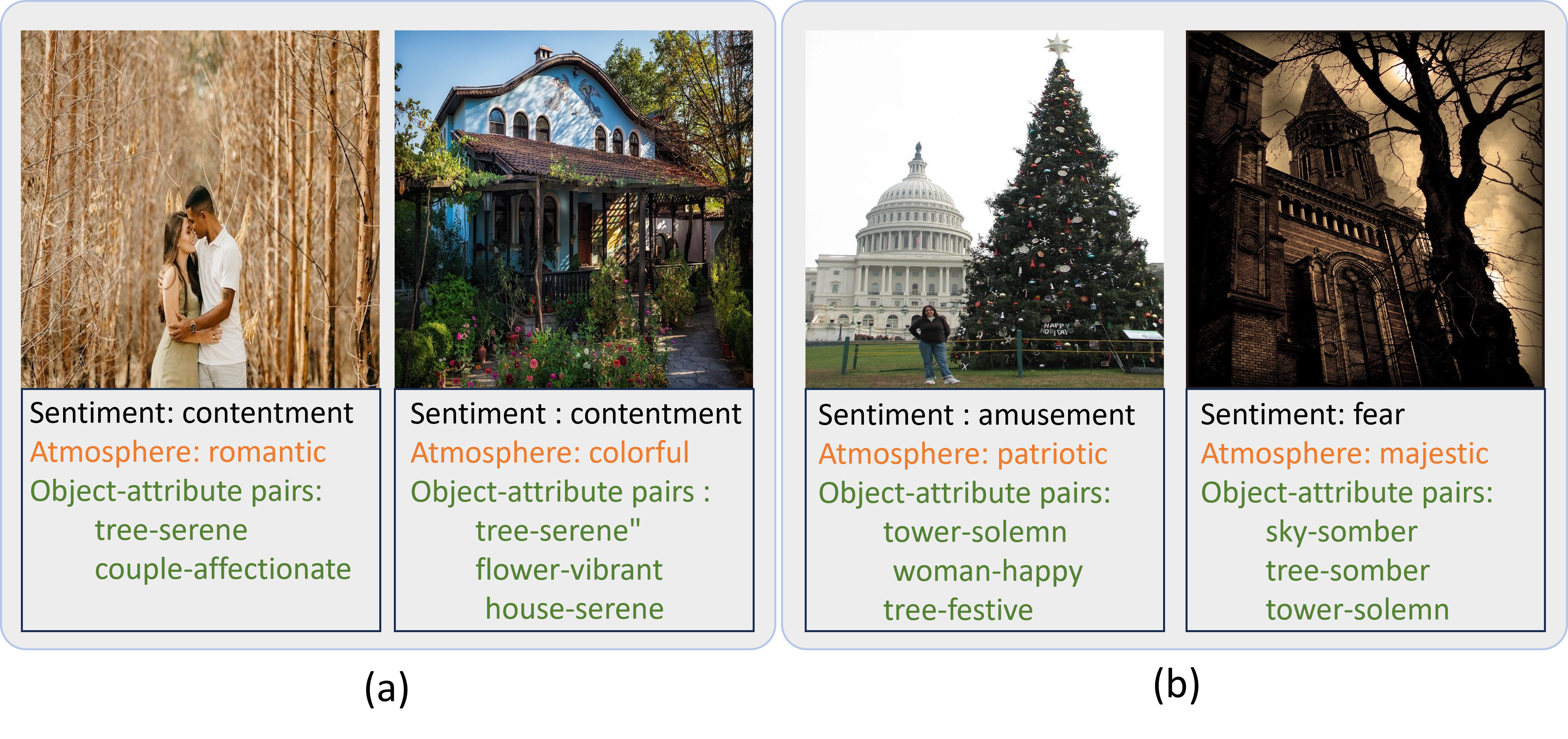}
  \caption{In (a), two images with different objects and corresponding attributes are categorized the same emotion ``contentment"; In (b), two images with different emotions have similar objects with different attributes.}
  \label{motivation}
\end{figure}


For exploring how to use text to alter images sentiment, early researchers proposed some models based on the relationships between emotions and colors \cite{yang2008automatic, peng2015mixed}, in which color features were selected according to the chosen emotion word and color histogram \cite{pouli2010progressive} were applied for image recoloring. Meanwhile, some studies leveraged the semantic connection between emotion words and images by retrieving suitable emotion images from emotion database, and transferred their texture and color features to the content image. However, relying solely on handcrafted mappings between emotion words and colors limits the model's ability to modify image sentiment effectively. Additionally, emotion database with limited capacity may not always contain suitable target images to guide sentiment transfer.

For establishing the relationships between image and sentiment and realizing more effective image sentiment transfer, some researchers proposed to construct emotion spaces as a bridge between text and images \cite{yang2024emoedit,weng2023affective,kang2023emogen}. The current methods of constructing emotion spaces are broadly categorized into two types. One is to construct emotion space based on manually constructed semantic features, and the other is to construct emotion space by using deep learning to obtain the mean and variance of different emotions in the feature space. The first one's performance is constrained by the semantic range and density covered by the handcrafted features, which struggled to capture abstract and complex emotions \cite{yang2024emoedit,weng2023affective}.  The second one represents the emotion space using only mean and variance may overlook more complex high-order statistical information, making it difficult to fully describe the true distribution of emotions in the feature space \cite{kang2023emogen}.

\IEEEpubidadjcol

To fully mine the emotion correlations between text semantic and image features, we try to construct an emotion space enriched with the semantic information for effective text-driven image sentiment transfer. Observations from a large-scale sentiment image dataset reveal that due to the complexity and diversity of emotions, images with the same emotion category can be expressed by different objects and attributes.  As shown in Figure. \ref{motivation}, under the same sentiment “contentment", the first image is constituted by affectionate couple and presents romantic atmosphere, and the second one is comprised by “serene house" and “vibrant flowers". The above detailed descriptive words can influence the emotion of different objects, ultimately affecting the overall emotion expression of an image. Similarly, images from different emotion categories may contain the same objects, but the descriptive words modifying these objects can vary, leading to significant differences in the overall conveyed emotion. For example, the third one and the fourth one have the same object “buildings", but different attributes and atmosphere make them have different emotions. These findings inspire us to leverage the complex interrelationships among emotion elements within images to construct emotion latent space, and use it to deep analyze the relationships between emotions, objects and attributes. 


Based on above analysis, we constructed a new Emotion Latent Space (EmoLat) in this paper, which stores the optimized latent distribution of emotion features that learned by adversarial training, enabling the emotion text to emotion image accurate mapping in image sentiment transfer task. In addition, we proposed a novel EmoLat based cross-modal sentiment transfer network, which leveraged the EmoLat and text semantic features as guidance and employed a multi-dimensional loss function to train a semantic mapper and a multi-modal Transformer, realizing accurate sentiment transfer from text to images. To construct EmoLat, we also constructed a new large-scale image sentiment dataset that contains semantically rich objects and attributes, Emospace Set, which consists of 118100 images, 1953 adjectives, and 27625 object-attribute pairs, which can be extensively used for various image sentiment analysis tasks. The contributions of this paper can be summarized as follows:

\begin{itemize}
    \item We constructed a novel Emotion Latent Space (EmoLat) by encoding optimized emotion representations to provide a unified and robust embedding space for more accurate cross-modal sentiment understanding and image emotion transfer.
    \item An EmoLat based cross-modal sentiment transfer network is proposed, which jointly exploits semantic and emotional cues, equipped with a multi-modal Transformer and a multi-dimensional training objective to ensure accurate and controllable image sentiment transfer.
    \item To construct EmoLat, we constructed a new large-scale image sentiment dataset Emospace Set, which offers fine-grained sentiment annotations with rich object–attribute pairs, enabling more comprehensive research in multimodal emotion analysis.
    \item Extensive experiments on large scale datasets demonstrate that our proposed EmoLat based cross-modal sentiment transfer network can achieve excellent performance in text-driven image eomotion transfer and significantly surpasses the state-of-the-art methods.
\end{itemize}

\section{RELATED WORK}

Text-driven image sentiment transfer is an emerging topic in computer vision, whose research results are still limited. In these section we will review some close related research works with text-driven image sentiment transfer, including: image sentiment analysis, image style transfer and image sentiment transfer. 

\subsection{Image Sentiment Analysis}

Image sentiment analysis focuses on analyzing various features of images to obtain the evoked emotions of image. Since object semantics has an important influence on the emotion evoked by images, some researchers try to realize image sentiment analysis by manually crafting object semantic features. For example, Borth et al. \cite{borth2013sentibank} designed a large-scale visual emotion ontology containing over 3,000 adjective-noun pairs (ANPs) SentiBank, followed by image emotion classification using CNNs \cite{chen2014deepsentibank}. 

With the rapid development of deep learning, researchers began to utilize deep neural networks to detect object semantic information in images for emotion analysis. Ali et al. \cite{ali2017high} selected specific object regions, extracted their deep features, and combined them with the full image to predict emotions. Zhang et al. \cite{zhang2016unconstrained} detected salient objects as local information to improve the accuracy of emotion prediction. Zhang et al. \cite{zhang2024object} proposed an object evoked emotion analysis network to study the interaction between objects and emotions for more accurate image emotion classification. 

Some researchers found that the relationships between objects played an important role in the study of image emotion analysis. Zhang et al. \cite{zhang2021graph} employed graph structures to represent the object semantics of an image and their relationships for more accurate emotion analysis. In addition, they further proposed a multi-level emotion region correlation analysis method to explore the multi-level correlation between object visual features and emotions.

\subsection{Image Style Transfer}

The task of image sentiment transfer is similar with image style transfer, and many methods are inspired by image style transfer methods. Here, we will introduce some text-guided image style transfer methods that close relate with our proposed methods. 


Text-guided style transfer uses text as the guidance for the style. Due to the significant feature differences between text and images, a cross-modal model is required to map the textual descriptions and the style features of the generated image into the same space. Kwon et al. \cite{kwon2022clipstyler} proposed the first text-guided style transfer model Clipstyler, which used CLIP to project text and images into the same space and guided style transfer. Shortly thereafter, Kim et al. \cite{kim2022diffusionclip} proposed DiffusionCLIP, which gradually generated images through a diffusion process, while optimizing the matching between the generated image and the text prompt by CLIP.  Liu et al. \cite{liu2023name} introduced a contrastive training strategy based on CLIP to effectively align stylized images with textual descriptions. Wu et al. \cite{wu2024textstyler} proposed STNet to maximize the retention of detailed information from the content image while performing style transfer.

\subsection{Image Sentiment Transfer}
Image sentiment transfer is an important branch of visual sentiment analysis that has garnered increasing attention in recent years. In the early, researchers focuses on the relationship between low-level features and emotions, modifying the color and texture features of an image according to the target emotion to alter the image's emotion tone. For example, Yang et al. \cite{yang2008automatic} proposed a color emotion transfer framework that established a mapping between color and emotion based on psychological theories. Xu et al. \cite{jin2013image} learned emotion-related knowledge from a set of target images with emotion annotations for image sentiment transfer to avoid human bias and subjectivity. Wang et al. \cite{wang2013affective} created a model that links emotion words with color themes, enabling users transfer emotion by modify image colors. Then, He et al. \cite{he2015image} proposed to provide color schemes based on a predefined color-sentiment model for image sentiment transfer. Ali et al. \cite{ali2017automatic} altered the sentiment of the source image by adjusting color-weighted histogram based on the reference image's color distribution.

In addition, researchers discovered that controlling the alignment of image content could better guide the transfer of images's entiment. Based on this, Kim et al. \cite{kim2016image} proposed an image semantic matching method to align target semantic regions and used a color transfer algorithm to change evoked emotions of images. Afterwards, Liu et al. \cite{liu2018emotional} integrated semantic information of images and proposed an image sentiment color transfer framework consisting of four neural networks for image sentiment transfer. Considering the influence of the object in an image on the emotion of the image, Chen et al. \cite{chen2020image} proposed an object-level image emotion transfer framework SentiGAN. Then, An et al. \cite{an2021global} designed an image retrieval algorithm based on the SSIM index, effectively establishing content relationships between the input and target images, thereby improving the performance of sentiment transfer.

For effectively exploring the correlations between emotion and image, Zhu et al. \cite{zhu2023emotional} proposed a emotion-generating adversarial network, which embeds images from different emotion categories into a shared neutral high-level content and low-level emotion feature space. Afterwords, They also \cite{zhu2023text} introduced text information as a supplement to the high-level semantic information in sentiment transfer and proposed a text-guided emotion-generating adversarial network (TGE-GAN). Weng et al. \cite{weng2023affective} completely removed the image guidance factors and used text as the sole guide for sentiment. They innovatively proposed an affective image filter, which is capable of understanding the visually abstract emotions in text and reflecting them as visually concrete images with appropriate colors and textures. 

\textbf{In summary}, although recent research in image sentiment transfer has made notable progress, several key challenges remain: 1) the lack of exploration into transferring emotion features represented by emotional words to images; 2) the absence of an effective emotion transfer mechanism between text and images. To address these issues, we construct an Emotion Latent space (EmoLat) that bridges text and image modalities, providing a cross-modal feature fusion platform for text-driven image sentiment transfer. To fully leverage and integrate the advantages of the EmoLat in connecting text and images, we further proposed a cross-modal sentiment transfer network, which uses a visual-sentiment-semantic multi-dimensional loss function to achieve precise text-driven image sentiment transfer based on EmoLat. In addition, we build a large-scale image sentiment dataset, Emospace Set, which contains rich emotion annotations and image content to support the construction of an effective EmoLat. Extensive experiments conducted on Emospace Set demonstrate that our method achieves superior performance in text-driven image sentiment transfer and outperforms existing state-of-the-art approaches.


\section{CONSTRUCTION OF EMOSPACE SET}
An unified and well-structured dataset is the important foundation of research on image emotion transfer. However, there is currently no existing dataset that can fully meet the requirements of this task. Existing emotion datasets are limited by data size, coarse-grained emotion categories, and insufficient emotion annotations, and cannot meet the needs of in-depth research on image emotion transfer. Recent emotion datasets like EmoSet \cite{yang2023emoset} attempted to explore the relationship between image sentiment and various attributes (e.g., objects, colors, and scene elements), they still fall short in several aspects: 1) Incomplete object annotations – EmoSet lacks fine-grained attribute annotations, limiting its utility for understanding the nuanced composition of images' sentiment; 2) Overly generalized  emotion labels –  compared with eight emotion categories in EmoSet, whereas real-world emotion experiences are more diverse, and images within the same emotion category usually exhibit varying visual contents; 3) Limited emotion expressiveness in visual features – EmoSet primarily categorizes images based on broad emotion labels, without incorporating the subtle variations in visual attributes that contribute to different emotion experiences.

To make up for the shortcomings of EmoSet, we create a more comprehensive dataset - Emospace Set based on EmoSet. Emospace Set not only provides a richer and more fine-grained representation of emotions but also includes a wider range of emotion attributes, supporting future research in image sentiment transfer more effectively. In addition, extensive manual filtering and validation were conducted to further enhance the quality and reliability of the dataset. All experiments in this work are conducted on Emospace Set, and the dataset will be made publicly available to facilitate future research.

The Emospace set is an expansion and enhancement of Emoset. We re-annotated the objects in the images of the dataset using the Tag2Text model \cite{huang2023tag2text}, and then removed duplicate nouns in the image labels by natural language processing techniques, obtaining a more refined set of objects. After that, we used the Llava model \cite{liu2024improved} to generate attribute annotations for the images. Specifically, we first applied a global attribute annotation prompt: “From a global perspective of the image, provide an adjective that can describe the image." Llava returned an appropriate global descriptive word based on the image. Then, for each object in the image, we constructed an object-specific attribute annotation prompt: “From the perspective of \{object\} in the image, provide a corresponding adjective that describes the emotion that the \{object\} conveys in the image." Llava generated attribute word of each object based on the context. Finally, we obtained global attribute annotations and object-specific attribute annotations of each image. The Emospace set contains 118100 images, with each image having up to 14 object-attribute pairs. The entire dataset includes total of 1953 attributes and 27625 unique object-attribute pairs, providing rich semantic and emotion information for the construction of the EmoLat. The Emospace set has been uploaded to http://github.com/JingVIPLab/EmoLat and is open to all researchers.

\section{METHODOLOGY}
In this section, we first introduce the construction of Emotion Latent space (EmoLat), then present the EmoLat based cross-modal sentiment transfer network, including its training framework and loss functions.

\subsection{Construction of Emotion Latent Space}
\begin{figure}[h]
  \centering
  \includegraphics[width=\linewidth]{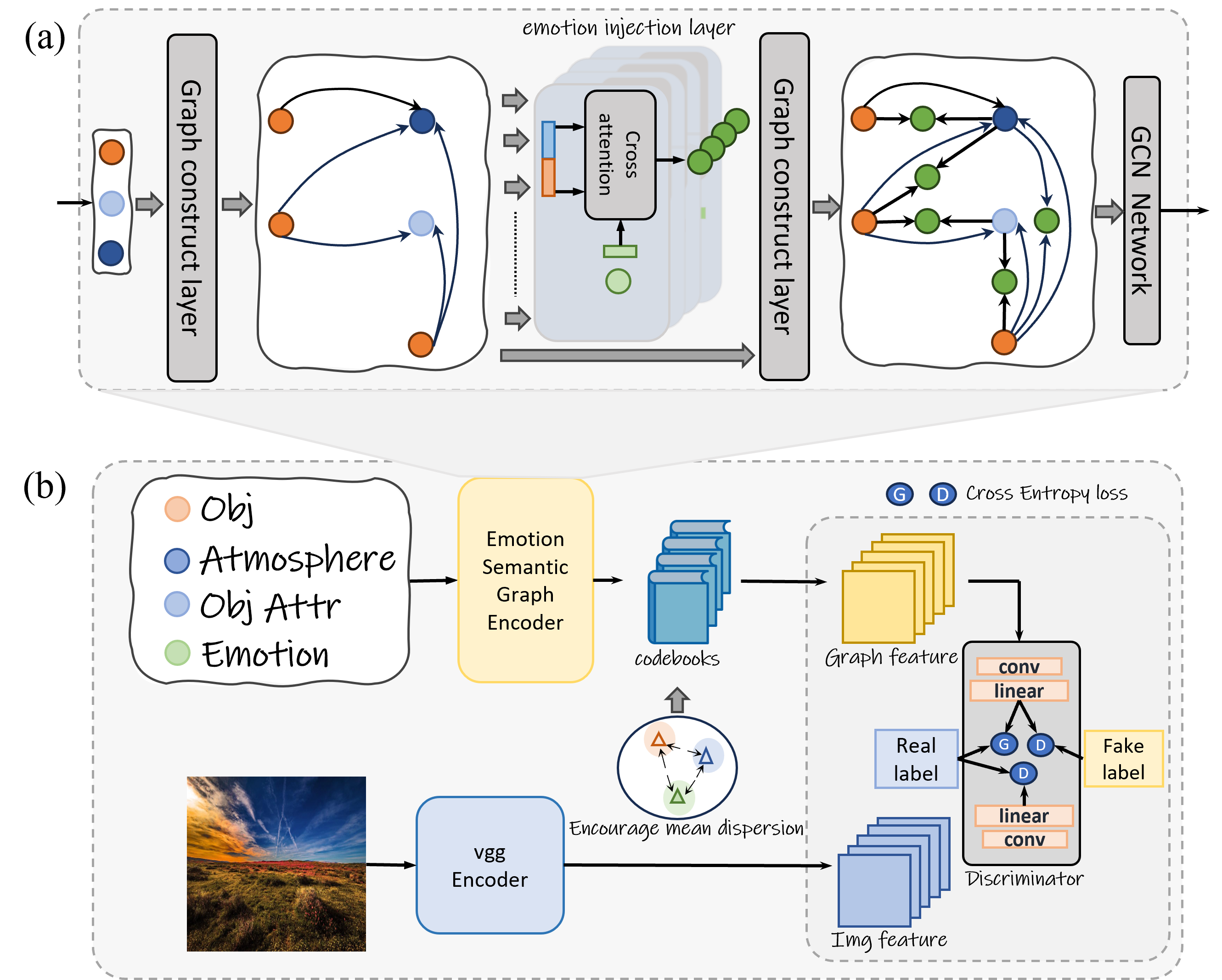}
  \caption{(a) The structure diagram of emotion semantic graph encoder. (b) The structure diagram of the emotion latent space generator is as follows: images are fed into a frozen VGG encoder to extract features, which serve as real data for the network. After extracting features from the emotion semantic map through the encoder, the data is processed by one of codebooks to obtain generated data. The discriminator achieves the feature distribution alignment between the generated data and real data by optimizing the loss function.}
  \label{ESGAN}
\end{figure}

For realizing more effective image sentiment transfer, we try to construct an Emotion Latent space (EmoLat) to present the complex and rich relationships between text semantic and image emotions, whose framework is illustrated in Figure. \ref{ESGAN}. 

First, we build an emotion semantic graph using emotions, objects, and attributes, whose encoder consists of two graph construction layers, an emotion semantic injection layer, and a graph convolutional network \cite{kipf2016semi, johnson2018image}. Then, we incorporate the training paradigm of a Generative Adversarial Network (GAN) \cite{goodfellow2014generative} to construct the EmoLat. Specifically, we extract image features from a pre-trained VGG \cite{simonyan2014very} as the source of real data, and the emotion semantic graph encoder generates synthetic data combined with vector quantization \cite{van2017neural}.  Then we train the discriminator to distinguish between real and generated data, and a mean dispersion incentive loss is used to ensure the rationality of the distribution. Next, we will introduce the whole process in detail.

\noindent\textbf{Emotion semantic graph encoder.} As shown in Figure 2(a), the first directed graph is constructed based on the object-attribute pairs for each image. We extract features ${f}_{O}$, ${f}_{A}$ and ${f}_{E}$ by CLIP's text encoder, which represent the features of objects, attributes, and emotions, respectively. These features are then input into the graph construction layer to generate the first directed graph. In this graph, objects serve as the starting nodes, while attributes serve as the ending nodes. Additionally, all object nodes are connected to a global adjective node. We use ${f}_{O}$, ${f}_{A}$ and ${f}_{E}$ to represent the object nodes, adjective nodes, and emotion nodes, respectively, and ${R}_{ij}$ to denote the relationships between each pair of nodes. The graph construction can be formulated as follows:

\begin{equation}
G^{k}_{1}=CS(f_{O}, R_{null}, f_{A})\quad i, j \in \mathit{Label_{k}}
\end{equation}
where CS represents the graph construction layer, $G^{k}_1$ denotes the first directed graph of the k-th image, $R_{ij}$ refers to the relationships between objects and attributes within each graph, and $Label_k$ represents the label of the k-th image. $R_{null}$ represents the the empty set of nodes' relationships.

Subsequently, we concatenate the node features of each directed edge and input them into the emotion semantic injection layer along with the emotion word features. Through a cross attention mechanism, we obtain the sentiment fused features. The formula is defined as follows:

\begin{equation}
f_{oa} = concat(f_O,f_A)
\end{equation}

\begin{equation}
Attn(Q,K,V) =Softmax(\left(\frac{QK^T}{\sqrt{d}}\right)V) \quad d \in \mathbb{R}
\end{equation}

\begin{equation}
f^{sem}_i = Attn({f}_{E},f_{oa},f_{oa})
\end{equation}
where $f^{sem}_i$ represents the sentiment fused features, and $Attn(Q,K,V)$ denotes the attention operation.

The $f^{sem}_i$ are used as nodes and are fed into the graph construction layer along with the first directed graph to generate the second emotion semantic graph. In this emotion semantic graph, each object and attribute is directed towards the node represented $f^{sem}_i$. The formula is defined as follows:

\begin{equation}
G^{k}_{2}=CS(G^{k}_{1}, R_{all}, f^{sem}_i)\quad i\in \mathit{Label_{k}}
\end{equation}

Finally, the second emotion semantic graph is fed into the GCN to obtain the final semantic graph features, which is defined as follows:

\begin{equation}
f_{graph}=GCN(G^{k}_{2})
\end{equation}

\noindent\textbf{Emotion latent space generator.} As shown in Figure 2(b), the GAN-based training approach enables the visual features of the images to serve as ground truth for achieving alignment between visual features and graph features, and guide the semantic graph feature distribution to be more reasonable.

We freeze all parameters of the pre-trained VGG encoder to extract image features, and use the output of the penultimate layer, $f_{img}$ as the input for real data. The emotion semantic graph encoder acts as the generator in the GAN framework, producing the initial graph features $f_{graph}$. Subsequently, we define the number of codebooks based on the same number of sentiment types in the Mikels' Wheel \cite{mikels2005emotional} of emotions. Vector quantization is then employed to replace the features in the codebooks, as defined by the following formula:

\begin{equation}
Q(z) = \underset{\mathbf{z}_{k} \in \mathcal{Z}}{\arg \min }\left\|z-\mathbf{z}_{k}\right\|
\end{equation}

\begin{equation}
f_{QG} = Q(f^i_{graph}) \quad i \in \{1, 2, \ldots, 8\}
\end{equation}
where, $z_k$ denotes the $k$-th entry in one codebook, $N$ is the number of entries in one codebook. $f_{QG}$ represents the features output by the codebook, and $i$ corresponds to one of the eight emotion types.

To prevent the feature distributions of each codebook from excessively overlapping in the emotion latent space, we employ a mean dispersion incentive loss to differentiate the mean of each feature cluster center. The formula is as follows:

\begin{equation}
\mu_i=\frac{1}{n_i}\sum_{k=1}^{n_i}f_{i,k}
\end{equation}

\begin{equation}
\mathcal{L}_{\mathrm{mdi}}= \frac{1}{C^2}\sum_{i=1}^C\sum_{j=1,j\neq i}\|\mu_i-\mu_j\|^2
\end{equation}
where $\mu_i$ represents the mean of the features in the i-th codebook, and $f_{i,k}$ denotes the k-th feature vector in the i-th codebook. $n_i$ indicates the number of features in the i-th codebook, while $C$ refers to the total number of codebooks. $\lVert \mu_i - \mu_j \rVert$ denotes the Euclidean distance between the means of the i-th and j-th codebook.

Subsequently, we feed the image features $f_{img}$ along with their corresponding real labels and the processed graph features $f_ {QG}$ with their corresponding fake labels into the discriminator. Here,  the real label is defined as a $1 \times 8$ vector $y_{real}$. Similarly, a fake label vector $y_{fake}$ is defined for the fake data. We replace the traditional Binary Cross-Entropy (BCE) loss in the discriminator with cross-entropy loss, as the latter can more effectively measure the difference between the predicted probability distribution and the true distribution, providing richer gradient information. Therefore, the loss function for the discriminator can be defined as follows:

\begin{equation}
\mathcal{L}_{real}^i = -\sum_{j=1}^C(y_{\mathrm{real}}\cdot\log D(f_{img})_{i,j}) 
\end{equation}

\begin{equation}
\mathcal{L}_{fake}^i = -\sum_{j=1}^C(y_{\mathrm{fake}}\cdot\log D(G(f_{QG}))_{i,j}) 
\end{equation}

\begin{equation}
L_D=\frac{1}{N}\sum_{i=1}^N (\mathcal{L}_{real}^i + \mathcal{L}_{fake}^i)
\end{equation}
where $D$ represents the discriminator, and $G$ denotes the generator, which is composed of the emotion semantic graph encoder and the codebooks. $\log D(f_{img})_{i,j}$ represents the logarithmic predicted probability of the i-th real sample in j-th emotion class by the discriminator. $\log D(G(f_{QG}))_{i,j}$ denotes the logarithmic predicted probability of the i-th generated sample in j-th emotion class. Both $y_{real}$ and $y_{fake}$ are $1 \times 8$ vectors. $C$ and $N$ represent the number of categories and batches respectively.

Finally, the $\mathcal{L}_{\mathrm{mdi}}$ is incorporated as part of the loss of generator. Thus, the loss function for training the generator in the emotion latent space generator is defined as:

\begin{equation}
L_G=-\frac{1}{N}\sum_{i=1}^N\sum_{j=1}^C\log D(G(f_ {QG}))_{i,j} - \mathcal{L}_{\mathrm{mdi}}
\end{equation}

\begin{figure}[!t]
  \centering
  \includegraphics[width=2.5in]{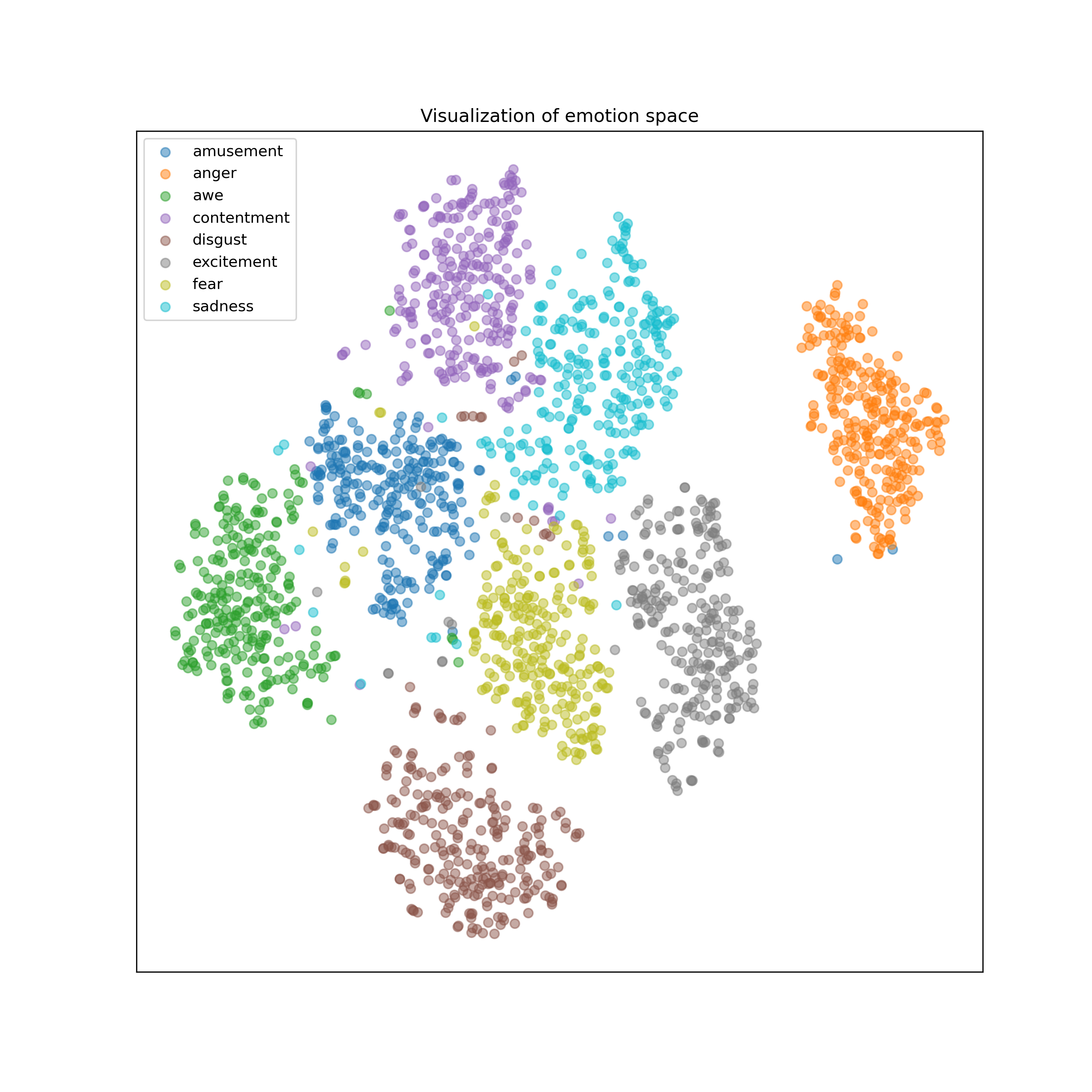}
  \caption{t-SNE Visualization of EmoLat.}
  \label{EmoLat}
\end{figure}

After multiple iterations, we visualized the emotion features stored in the codebooks, as shown in Figure. \ref{EmoLat}. Different colors represent the feature distribution of different emotions. It can be intuitively observed that, under the guidance of the image as the groundtruth, the distribution of emotion semantic features from emotion semantic graph gradually becomes reasonable, and the feature differences between different emotions are obvious.

\begin{figure*}[!t]
  \centering
  \includegraphics[width=\linewidth]{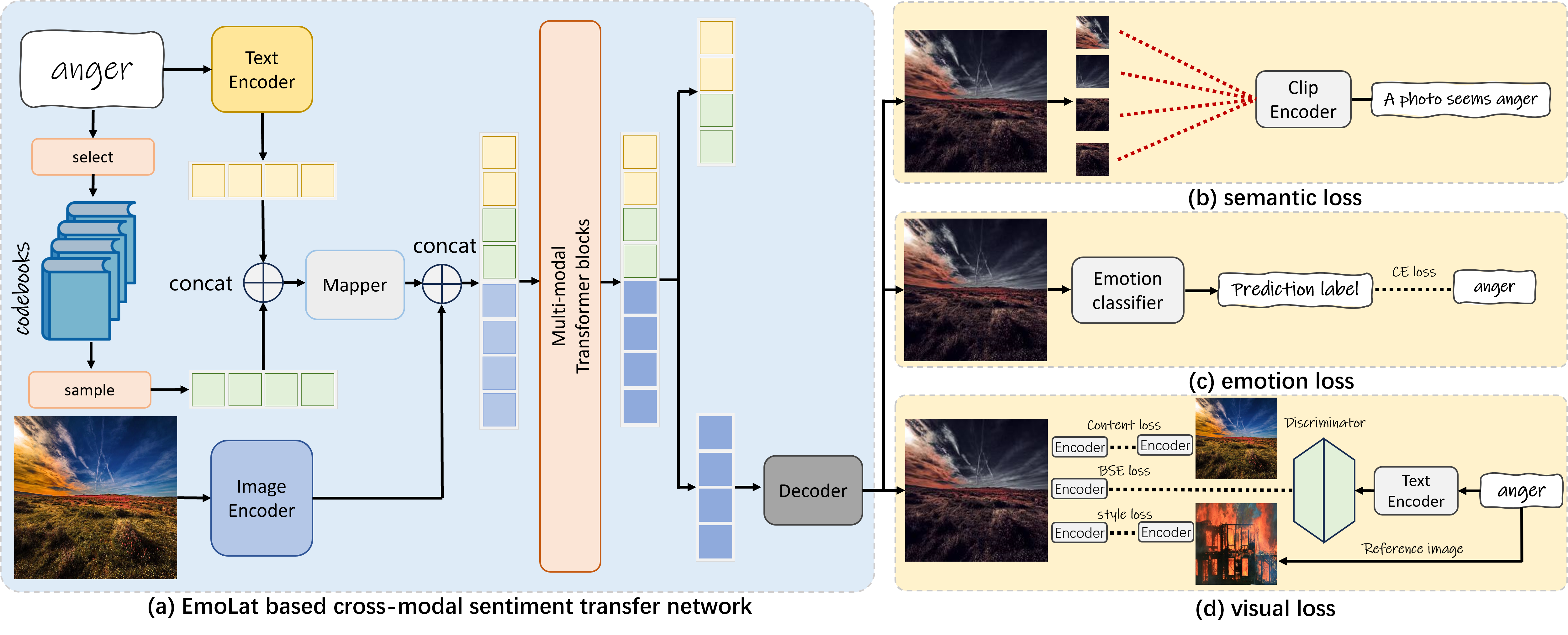}
  \caption{EmoLat based cross-modal sentiment transfer network training framework.}
  \label{network}
\end{figure*}

\subsection{EmoLat based cross-modal sentiment transfer network} 
After constructing the EmoLat, we obtained codebooks that stored the emotion features. Next, we will introduce the EmoLat based cross-modal sentiment transfer network and the multi-dimensional loss functions in the training process.

As shown in Figure. \ref{network}, the EmoLat based cross-modal sentiment transfer network has two types of inputs: text and images. First, the text is fed into the CLIP text encoder to extract semantic features $f_e$. And the corresponding emotion type is selected from the codebooks, and random sampling is performed to output emotion features $f_c$. The semantic and emotion features are then concatenated to obtain $f_{ec}$. The $f_{ec}$ is passed into the feature mapping layer to produce emotion-semantic features $f_m$, which can be expressed by the following formula:

\begin{equation}
f_m = Mapper(f_{ec})
\end{equation}

Meanwhile, the image is fed into the frozen convolutional encoder to extract visual features $f_{img}$. The emotion-semantic features and visual features are then fed into a multi-modal Transformer \cite{weng2023affective}. The process can be expressed by the following formula:

\begin{equation}
[\bar{f}_{m,i}]=\mathrm{MSA}(\mathrm{LN}([f_{m,i-1}]))+[f_{m,i-1}],  i\in\{1,\ldots,L\}
\end{equation}

\begin{equation}
[f_{fuse}]=\mathrm{MLP}(\mathrm{LN}([\bar{f}_{m,i}]))+[\bar{f}_{m,i}],  i\in\{1,\ldots,L\}
\end{equation}

Finally, the output features are separated according to the concatenation order used previously. The part corresponding to the image features is fed into the image decoder for image reconstruction. The process can be expressed by the following formula:

\begin{equation}
\Hat{f}_{img}, \Hat{f}_m = split(f_{fuse})
\end{equation}

\begin{equation}
I_{img} = decoder(\Hat{f}_{img})
\end{equation}

To improve the performance of EmoLat based cross-modal sentiment transfer network, we optimize the model using three different types of loss functions, including visual loss, emotion loss, and semantic loss. The visual loss minimizes image content degradation while enhancing the similarity between the generated image’s visual style and the style of images corresponding to the target sentiment. The emotion loss leverages a pre-trained emotion classifier to measure the discrepancy between the generated image and the target sentiment, ensuring that the transformed image aligns with the expected emotion characteristics. The semantic loss calculates the semantic distance between the generated image and the target sentiment in the CLIP space to guarantee  the correction of emotion transfer.

For the visual loss, we primarily calculate the content loss between the generated image and the content image, as well as the style loss and identity loss with respect to the reference image. Additionally, the GAN loss from \cite{weng2023affective} is incorporated as part of the visual loss.

The content loss $\mathcal{L}_{content}$ can be defined as follows:
\begin{equation}
\mathcal{L}_{content}=||{f}_{fuse}^{img}-f_{ori}||_2
\end{equation}
where $f_{fuse}^{img}$ is the generative images features of fixed decoder output. $f_{ori}$ represents the feature of content images.

The style loss $\mathcal{L}_{style}$ can be defined as follows:
\begin{equation}
\mathcal{L}_{style}=||\mu({f}_{fuse}^{img})-\mu(f_{style})||_2 + ||\sigma({f}_{fuse}^{img})-\sigma(f_{style})||_2
\end{equation}
where $\mu$ and $\sigma$ are the mean and variance functions, respectively.

The identity loss $\mathcal{L}_{id}$ can be defined as follows:
\begin{equation}
\begin{aligned}
\mathcal{L}_{id} & = ||(I_{ss}-I_{style})||_2 + \gamma||f_{ss}-f_{style}||_2
\end{aligned}
\end{equation}
where $I_{ss}$ and $I_{style}$ are the identity images and the sample texts corresponding to the reference images, respectively. And $f_{ss}$ and $f_{style}$ are their features respective. $\gamma$ = 0.01 is a hyper-parameter.

The calculation formula for the GAN loss is defined as follows:
\begin{equation}
\begin{aligned}
\mathcal{L}_{\mathrm{GAN}} &= \log D(f_{style}) + \log\left(1-D(G(f^{img}_{fuse}, f_{tex}))\right) \\
& + \log D(f_{style}, f_{tex}) + \log\left(1-D(G(f^{img}_{fuse}, f_{tex}), f_{tex})\right)
\end{aligned}
\end{equation}
where $f_{tex}$ represents the features of input simple text.

Therefore, the total visual loss can be defined as follows:

\begin{equation}
\mathcal{L}_{vis} = \lambda_1\mathcal{L}_{content} + \lambda_2\mathcal{L}_{style} + \lambda_3\mathcal{L}_{id} + \lambda_4\mathcal{L}_{\mathrm{GAN}}
\end{equation}

For the emotion loss, we use a pre-trained emotion classifier to compute the cross-entropy loss between the generated image and the label corresponding to the text, which is defined as follows:

\begin{equation}
\mathcal{L}_{emo} = -\frac{1}{N}\sum_{i=1}^{N}\sum_{c=1}^{C}y_{i,c}log(\hat{y}_{i,c})
\end{equation}

For the semantic loss, we introduce the patch loss from Clipstyler \cite{kwon2022clipstyler}, and use a pre-trained CLIP model to compute the semantic similarity $y_{clip(text, img)}$ between the text and the generated image. Therefore, the formula for the semantic loss can be defined as follows:

\begin{equation}
\mathcal{L}_{clip} = 1 - y_{patch(text, img)}
\end{equation}

The final total loss can be summarized as follows:
\begin{equation}
\mathcal{L}_{total} = \mathcal{L}_{vis} + \lambda_{emo}\mathcal{L}_{emo} + \lambda_{clip}\mathcal{L}_{clip}
\end{equation}
where $\lambda_{emo}$ and $\lambda_{clip}$ are two different hyper-parameters.

\section{EXPERIMENTS}
To validate the effectiveness of EmoLat based cross-modal sentiment transfer network, we conducted a series of experiments on our constructed Emospace Set. Next, we will introduce experimental settings, evaluation metrics, comparative experiments, ablation studies and qualitative analysis in detail.

\subsection{Experimental Settings and Evaluation Metrics}
To construct the EmoLat, all images in the Emospace Set are used to train the emotion latent space generator. For training the EmoLat based cross-modal sentiment transfer network, the Emospace Set is divided into eight emotion categories. In each category, the first 95\% of the images are used for training, while the remaining 5\% are used for testing.

The proposed methods are implemented by PyTorch framework \cite{paszke2017automatic}. When training the emotion latent space generator, we used AdamW \cite{loshchilov2017decoupled} as the optimizer for the emotion semantic graph encoder and codebooks. The batch size was set to 64, the learning rate was 5e-4, and the training spanned 3 epochs. For training the EmoLat based cross-modal sentiment transfer network, we used the Adam optimizer \cite{diederik2014adam}, in which the batch size was set to 4, the learning rate was 5e-4, and the training ran for 80000 iterations. All the models were trained on an NVIDIA GeForce RTX 3090 24GB GPU, requiring approximately 12 hours in total.

To evaluate the performance of the proposed method, we assessed it from two aspects: image sentiment transfer rate and image content quality. For the transfer rate, we used a pre-trained emotion classifier to determine whether the image sentiment after transformation matches the target emotion, with accuracy (Acc) representing the performance. For image content quality, we calculated the structural similarity of the grayscale images between the generated image and the content image using SSIM \cite{wang2002universal}, measured the pixel similarity between the generated image and the content image using Reconstructed error \cite{zhu2023emotional}, and evaluated the difference between the generated image distribution and the real image distribution using FID \cite{wang2004image}.

\begin{figure*}[!t]
\centering
\begin{tabularx}{\textwidth}{*{8}{>{\centering\arraybackslash}X}}
content & target & Ours & AIF & Ldast & TxST & SpectralCLIP & ClipStyler \\
\end{tabularx}
\centering

\subfloat{\includegraphics[width=0.85in]
{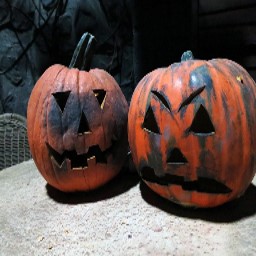}}
\hfil
\subfloat{\includegraphics[width=0.85in]
{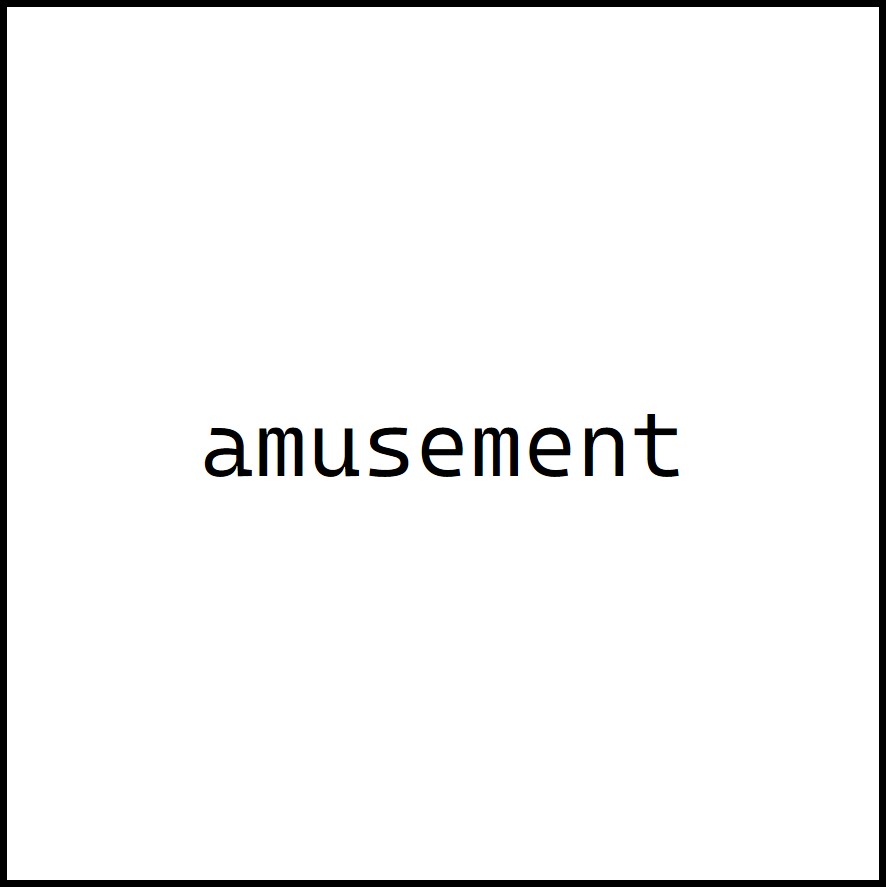}}
\hfil
\subfloat{\includegraphics[width=0.85in]
{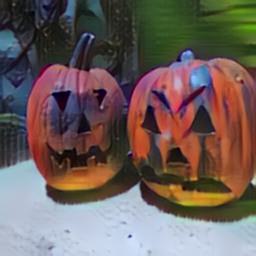}}
\hfil
\subfloat{\includegraphics[width=0.85in]
{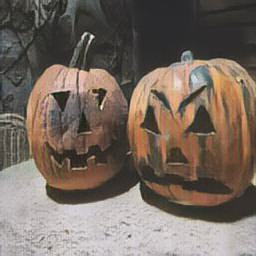}}
\hfil
\subfloat{\includegraphics[width=0.85in]
{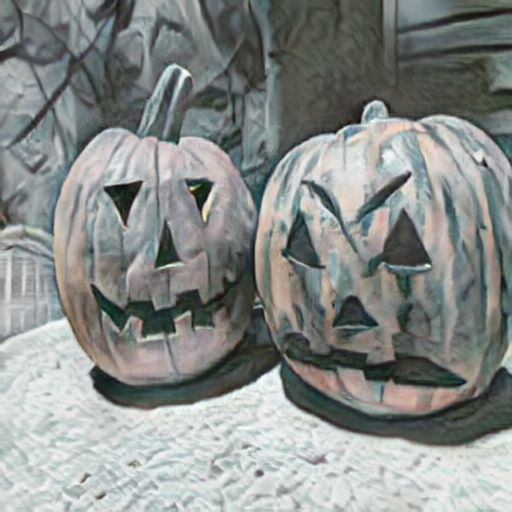}}
\hfil
\subfloat{\includegraphics[width=0.85in]
{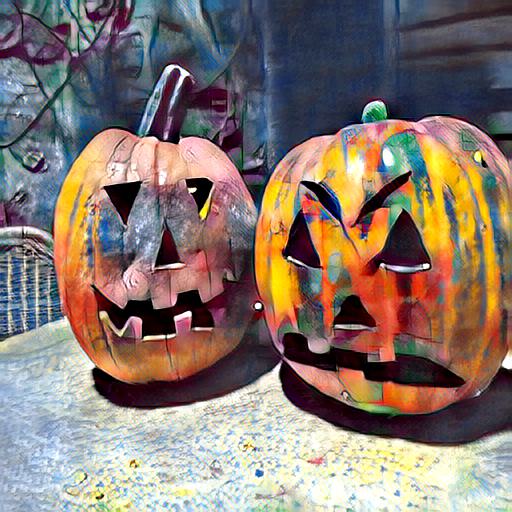}}
\hfil
\subfloat{\includegraphics[width=0.85in]{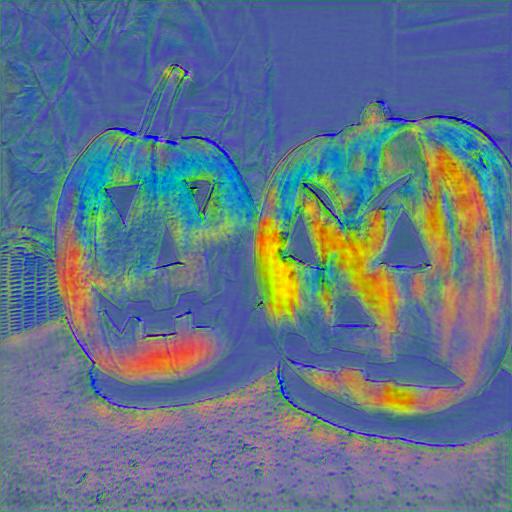}}
\hfil
\subfloat{\includegraphics[width=0.85in]{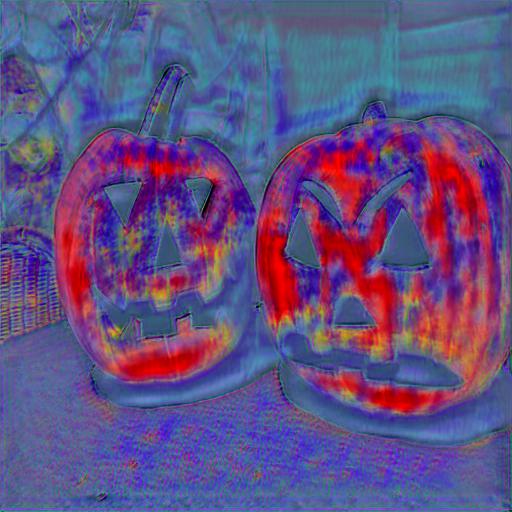}}
\vspace{4pt}

\vspace{-13pt}
\subfloat{\includegraphics[width=0.85in]
{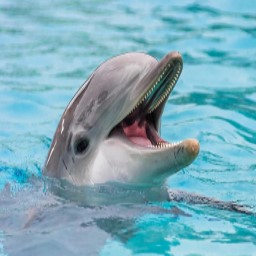}}
\hfil
\subfloat{\includegraphics[width=0.85in]
{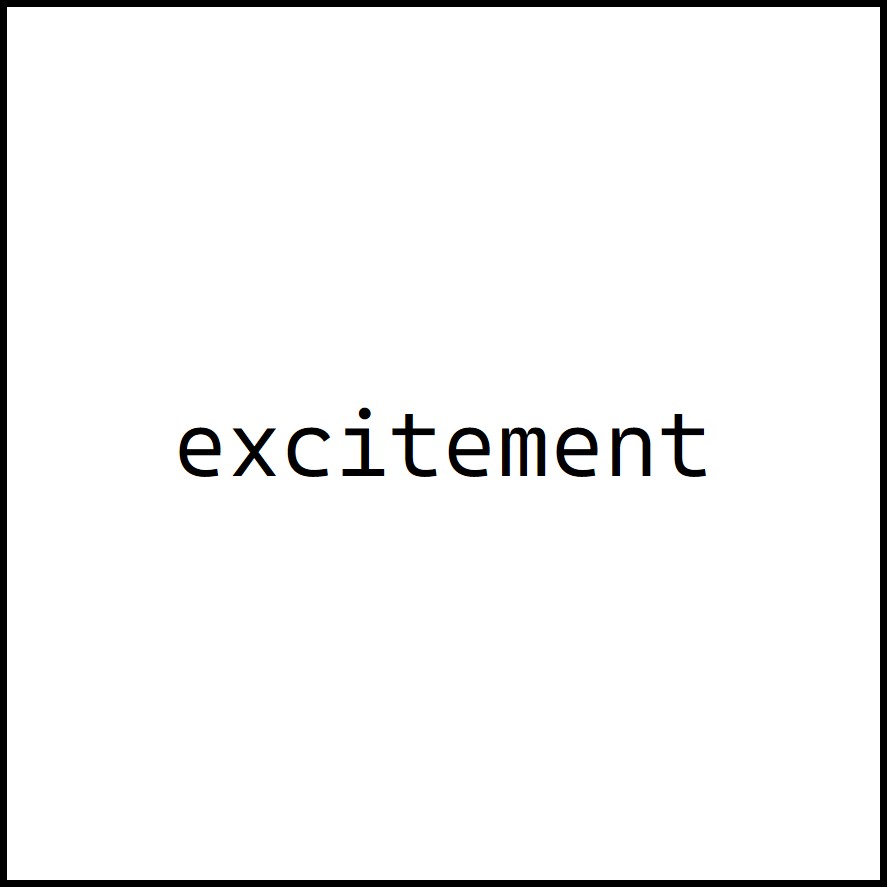}}
\hfil
\subfloat{\includegraphics[width=0.85in]
{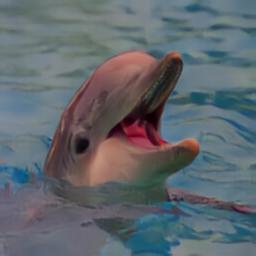}}
\hfil
\subfloat{\includegraphics[width=0.85in]
{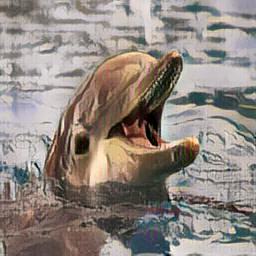}}
\hfil
\subfloat{\includegraphics[width=0.85in]
{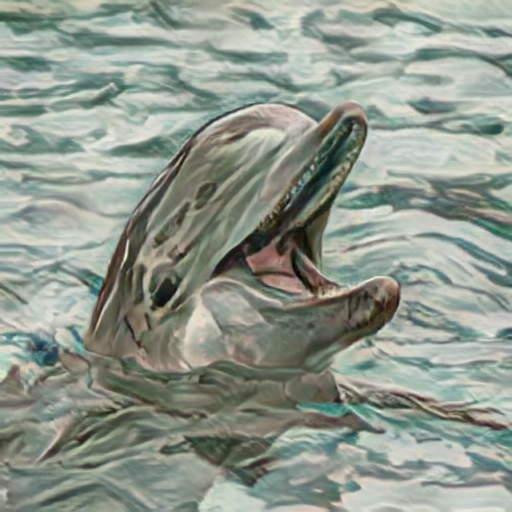}}
\hfil
\subfloat{\includegraphics[width=0.85in]
{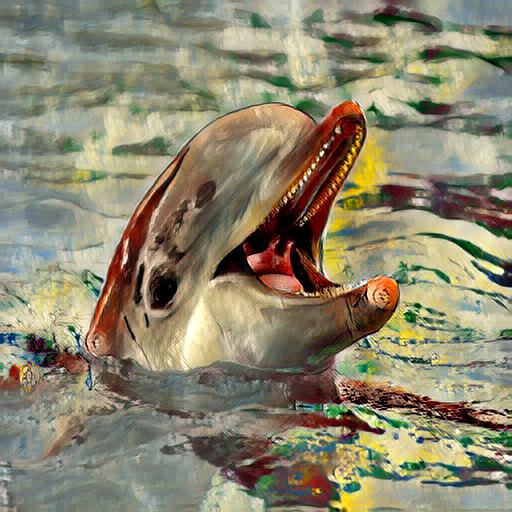}}
\hfil
\subfloat{\includegraphics[width=0.85in]
{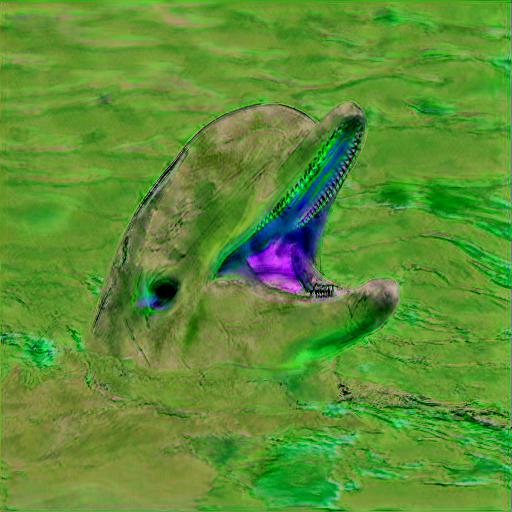}}
\hfil
\subfloat{\includegraphics[width=0.85in]
{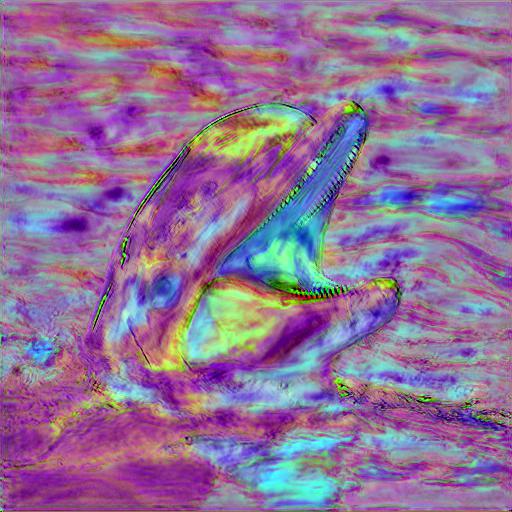}}
\vspace{4pt}

\vspace{-13pt}
\subfloat{\includegraphics[width=0.85in]
{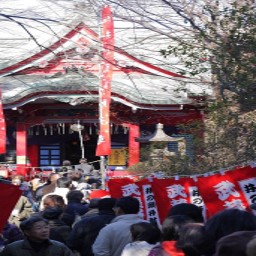}}
\hfil
\subfloat{\includegraphics[width=0.85in]
{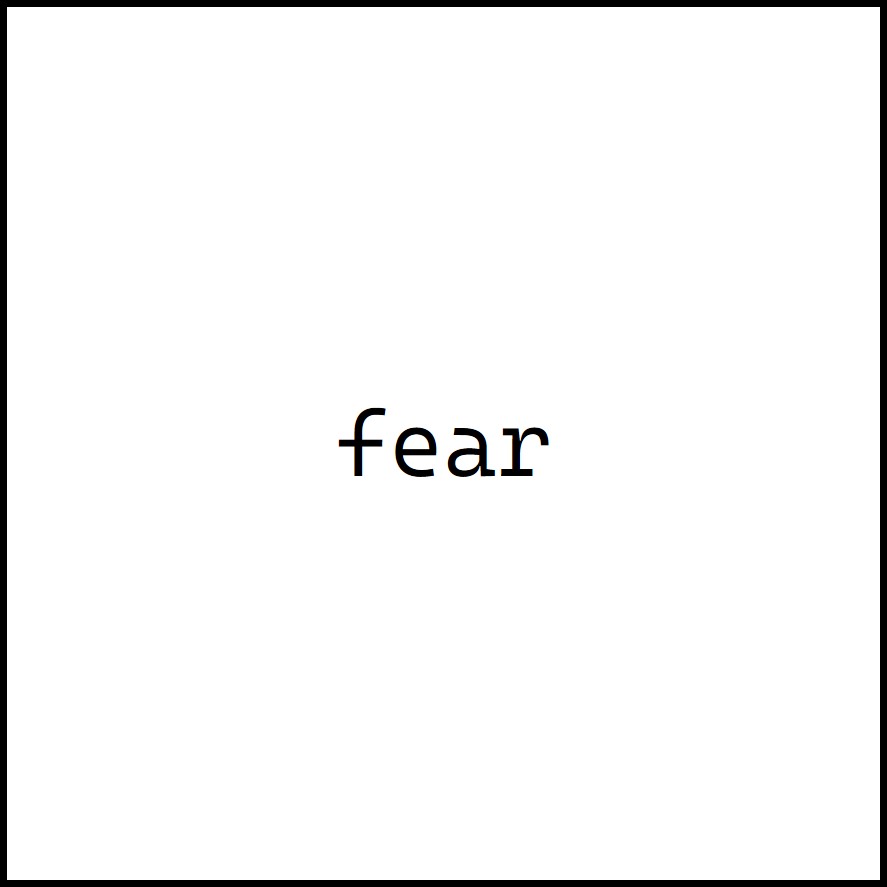}}
\hfil
\subfloat{\includegraphics[width=0.85in]
{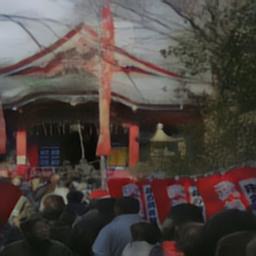}}
\hfil
\subfloat{\includegraphics[width=0.85in]
{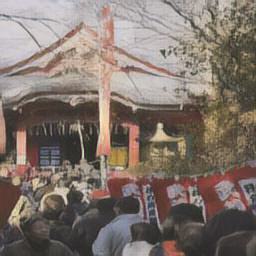}}
\hfil
\subfloat{\includegraphics[width=0.85in]
{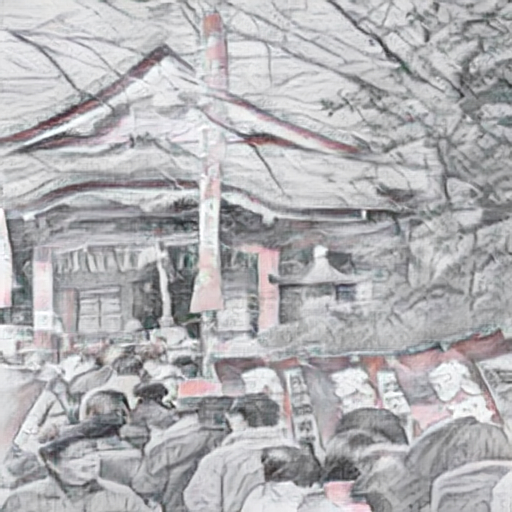}}
\hfil
\subfloat{\includegraphics[width=0.85in]{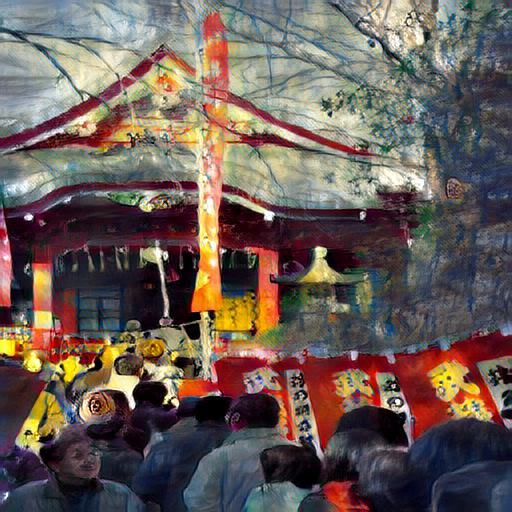}}
\hfil
\subfloat{\includegraphics[width=0.85in]
{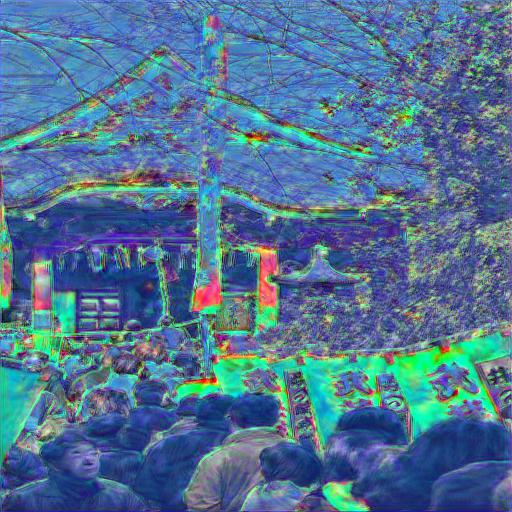}}
\hfil
\subfloat{\includegraphics[width=0.85in]
{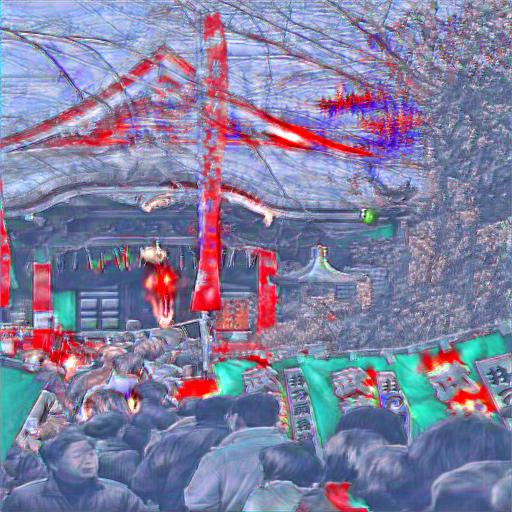}}
\vspace{4pt}

\vspace{-13pt}
\subfloat{\includegraphics[width=0.85in]
{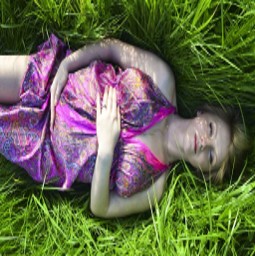}}
\hfil
\subfloat{\includegraphics[width=0.85in]
{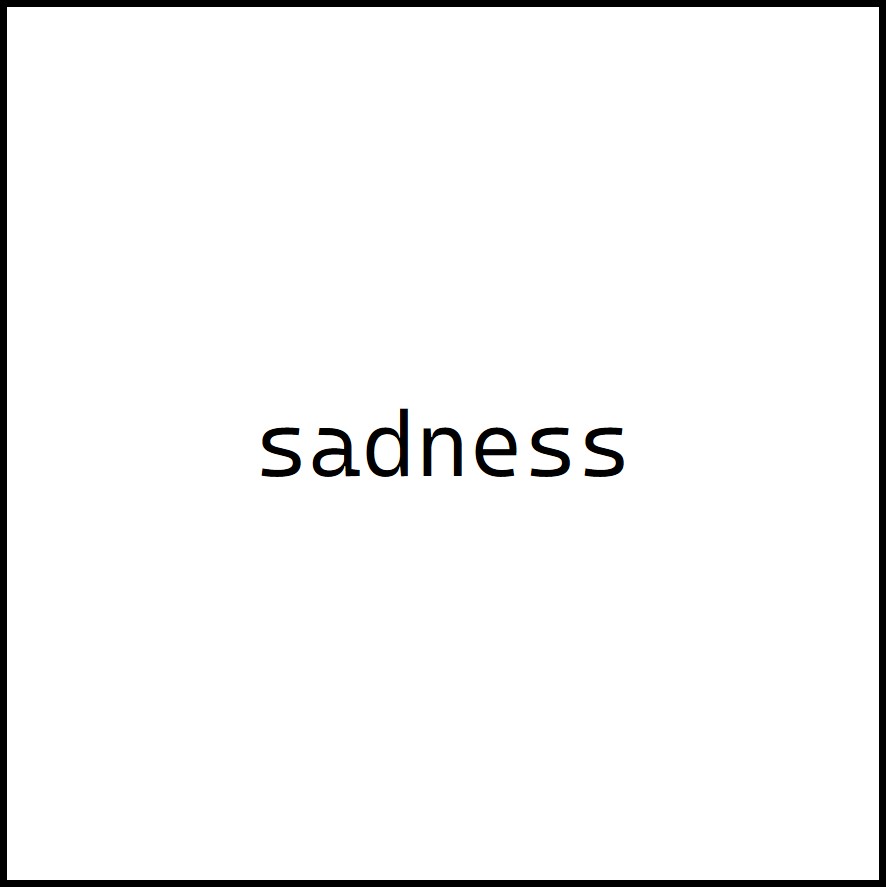}}
\hfil
\subfloat{\includegraphics[width=0.85in]
{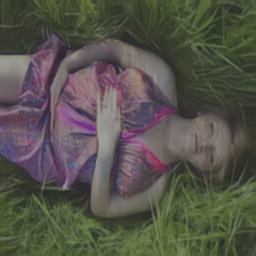}}
\hfil
\subfloat{\includegraphics[width=0.85in]
{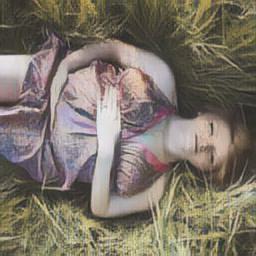}}
\hfil
\subfloat{\includegraphics[width=0.85in]
{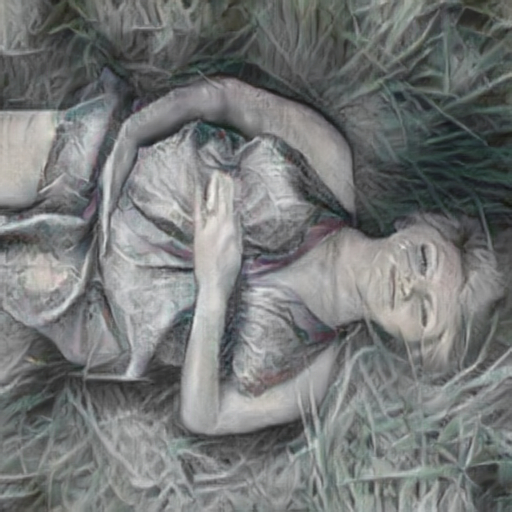}}
\hfil
\subfloat{\includegraphics[width=0.85in]
{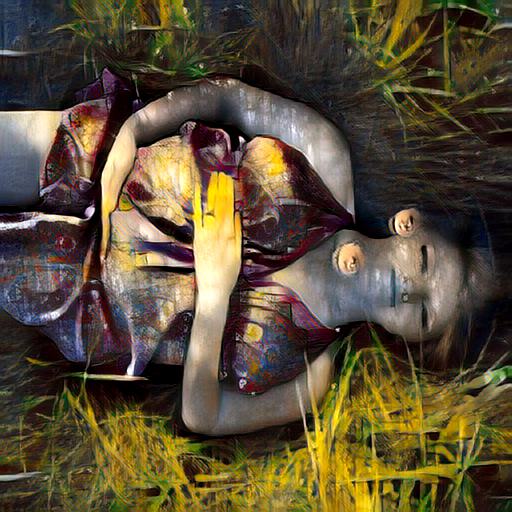}}
\hfil
\subfloat{\includegraphics[width=0.85in]
{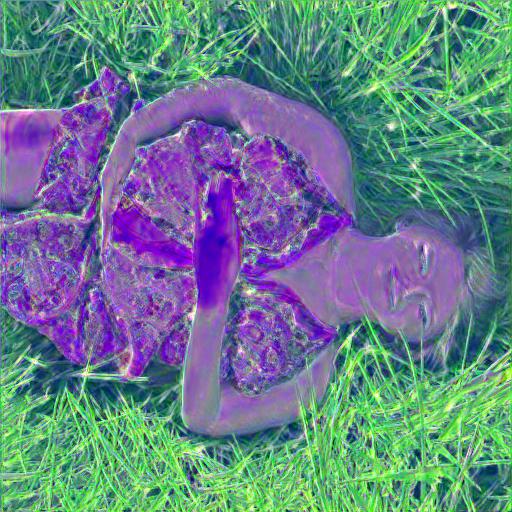}}
\hfil
\subfloat{\includegraphics[width=0.85in]{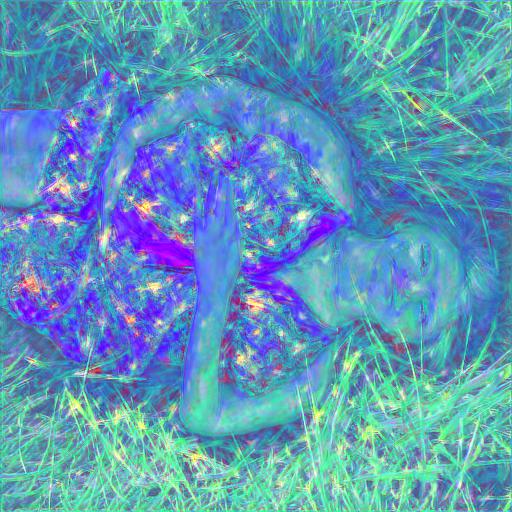}}
\caption{Qualitative Comparison of Different Sentiment Transfer Methods.}
\label{qualitative}
\end{figure*}

\subsection{Comparative Experiments} 
To comprehensively evaluate the effectiveness of our proposed text-driven image emotion transfer method, we conducted extensive comparative experiments on the Emospace Set. The research on text-driven image sentiment transfer methods is relatively scarce, only one method \cite{weng2023affective} has been published so far. Hence, we reproduced this work \cite{weng2023affective} and four advanced text-driven image style transfer methods \cite{kwon2022clipstyler, xu2024spectralclip, liu2023name, fu2021language}, and applied them to image sentiment transfer for comparative experiments. The experimental results are shown in Table \ref{tab:freq} and Table \ref{tab:freq1}

\begin{table}
\begin{center}
  \caption{Accuracy Comparison for Text-Driven Sentiment Transfer.}
\label{tab:freq}
\begin{tabular}{p{3cm}p{2cm}p{2cm}}
\hline
Method & Acc-8 & Acc-2 \\
\hline
Clipstyler \cite{kwon2022clipstyler} & 0.1339 & \textbf{0.5051} \\
SpectralCLIP \cite{xu2024spectralclip} & 0.1152 & 0.4801 \\
TxST \cite{liu2023name} & \underline{0.1931} & 0.3878 \\
Ldast \cite{fu2021language} & 0.1423 & 0.4047 \\
AIF \cite{weng2023affective} & 0.1126 & 0.3971 \\
\hline
Ours & \textbf{0.2337} & \underline{0.4944} \\
\hline
\end{tabular}
\end{center}
\end{table}

As shown in Table \ref{tab:freq}, in the comparison of sentiment transfer accuracy, our method achieved a sentiment transfer accuracy of 23.37\% in the eight-class sentiment transfer task, outperforming all compared methods. In the two-class sentiment transfer task, our method is only 1.07 percentage points lower than the current SOTA Clipstyler \cite{kwon2022clipstyler}.

Compared to AIF \cite{weng2023affective}, the only text-based image sentiment transfer method, our method improved sentiment transfer accuracy by 12.11\% in the eight-class task and 9.73\% in the two-class task. It showed that our EmoLat based cross-modal sentiment transfer network can accurately transfer target emotion to images, which also proved our proposed EmoLat can accurately present the emotion distribution of visual features, effectively distinguish different categories of emotion features. In addition, compared to emotion information from VAD dictionary, the emotion features provided by EmoLat contained richer emotion implicit information. 

From semantic perspective, emotion words can be considered a simple form of style description, meaning that image style transfer methods can realize sentiment transfer by handling style-like emotion features. Our reproduced four style transfer methods all used CLIP-based loss functions as semantic guidance for emotion transfer. Ldast \cite{fu2021language} extracted visual semantics from style descriptions using contrastive inference, but it struggled to capture sentiment-related style features when dealing with abstract emotion words. TxST \cite{liu2023name} introduced a decoder fusion module based on channel attention to optimize the selection and weighting of style features. SpectralCLIP \cite{xu2024spectralclip} and Clipstyler \cite{kwon2022clipstyler} employed basic PatchCLIP loss and Directional CLIP loss to control the direction of style transformation. In the experiments on Emospace Set, our proposed method achieved an average improvement of 8.76\% and 5\% in sentiment transfer accuracy over above four style transfer models in the eight-class and two-class tasks, respectively. The main reason is that our proposed method incorporated EmoLat as guidance during the encoding stage, providing more target sentiment guidance than the CLIP model based on specific semantics. Additionally, our method designed a visual-emotion-semantic joint loss during training process, which ensured that the image still maintained a high degree of emotion and content consistency after emotion transfer.

\begin{table}[h]
  \caption{Quantitative Comparison of Image Content Quality.}
  \label{tab:freq1}
  \begin{tabular}{lccl}
    \hline
    Method & SSIM $\uparrow$ & Reconstructed error $\downarrow$ & FID $\downarrow$ \\
    \hline
    Clipstyler \cite{kwon2022clipstyler} & 0.5051 & 56.08 & 192.74 \\
    SpectralCLIP \cite{xu2024spectralclip} & 0.4156 & 53.09 & 148.45 \\
    TxST \cite{liu2023name} & 0.5309 & 34.14 & 132.51 \\
    ldast \cite{fu2021language} & 0.4650 & 44.17 & 153.65 \\
    AIF \cite{weng2023affective} & \underline{0.6090} & \textbf{29.05} & \underline{115.27} \\
    \hline
    Ours & \textbf{0.6154} & \underline{31.59} & \textbf{86.69} \\
  \hline
\end{tabular}
\end{table}

As shown in Table \ref{tab:freq1}, our method achieved the highest scores in the comparison of image content quality by SSIM and FID, with values of 0.6154 and 86.69, respectively. In the Reconstructed Error, it ranked second with a score of 31.59 and was just lower than AIF \cite{weng2023affective} 2.54. Our method performs similarly to AIF on image quality, since both of them used visual loss functions to constrain the visual quality of generated images. 

Ldast \cite{fu2021language} applied GAN-based reconstruction method, which caused some content loss during image generation. TxST \cite{liu2023name} applied multi-scale decoder to capture content features at different levels. SpectralCLIP \cite{xu2024spectralclip} and Clipstyler \cite{kwon2022clipstyler} used a Unet-based architecture as the backbone for style transfer, but the sampling operations in the Unet structure led to some degree of content loss. Compared to above four CLIP-loss-based style transfer methods, our method improved the SSIM score by an average of 0.1362 and reduced the Reconstructed error and FID scores by an average of 11.72 and 61.83, respectively. This demonstrated that our proposed multidimensional loss function effectively enhanced the model’s ability to preserve image content, and was able to better preserve the content of the image while transferring emotion. 



\subsection{Ablation Studies}
To verify the effectiveness of the proposed EmoLat in image sentiment tranfer, we conducted a set of ablation experiment, in which the codebooks representing EmoLat is deleted from the entire framework, directly input the CLIP features of the text into the Mapper layer, and kept all other parameters unchanged.  

The ablation experimental results were shown in Table \ref{tab:freq2}. Without EmoLat, the model only achieved an accuracy of 43.51\% in the sentiment two-class classification and 13.78\% in the sentiment eight-class classification. Compared with the model with EmoLat, the accuracies decreased by 5.93\% and 9.59\% in the two-class and eight-class classification tasks, respectively. This demonstrated that EmoLat can effectively guide the image sentiment transfer model to generate images that align with the target sentiment.

\begin{table}[h]
  \caption{Ablation Results on Sentiment Transfer Accuracy.}
  \label{tab:freq2}
  \begin{tabular}{p{3cm}p{2cm}p{2cm}}
    \hline
    Method & Acc-8 & Acc-2 \\
    \hline
    w/o EmoLat & 0.1378 & 0.4351 \\
    Ours & 0.2337 & 0.4944 \\
  \hline
\end{tabular}
\end{table}

\subsection{Qualitative analysis}
We also conducted the qualitative analysis of the proposed method and five comparison methods, and the experimental results are shown in Figure \ref{qualitative}. The first column displayed the original images, the second column showed the target sentiment, the third column presented the generated images from our method, and the fourth to eighth columns showed the generated images from AIF \cite{weng2023affective}, Ldast \cite{fu2021language}, TxST \cite{liu2023name}, SpectralCLIP \cite{xu2024spectralclip}, and Clipstyler \cite{kwon2022clipstyler}, respectively.  

Obviously, the images generated by our proposed model are visually closer to real-world perceptions and align with the sentiment corresponding to the target text. This indicated that the EmoLat and the multidimensional loss function used for training the cross-modal sentiment transfer network not only improved sentiment transfer accuracy but also enhanced the overall visual quality of the images.  AIF, primarily trained on long text, struggled to capture sufficient semantic information when short text was used as a reference, even with simple VAD \cite{mohammad2018obtaining} features injected. As a result, its generated images often exhibited a single, monotonous style in terms of visual effect. Due to SpectralCLIP and Clipstyler did not have ability to understand abstract emotion words, their generated images lacked a real-world appearance. Although Ldast and TxST generated somewhat natural and realistic images. Since these two models do not process the emotion information, the emotions represented by the text are not effectively reflected in the generated images of them.

\section{CONCLUSION}
In this paper, we innovatively constructed an Emotion Latent space (EmoLat) and proposed an EmoLat based cross-modal sentiment transfer network to address the text-driven image sentiment transfer. To build a reasonable and effective EmoLat, we constructed a new large scale emotion dataset, Emospace Set, which enriches the object labels of images and provides rich attribute descriptions for the entire image as well as each object. Based on Emospace Set, we constructed EmoLat by a novel emotion latent space generator. At the same time, we proposed an EmoLat based cross-modal sentiment transfer network to achieve accurate text-driven image sentiment transfer. Experiments on Emospace Set proved that our proposed EmoLat based cross-modal sentiment transfer network can generate aesthetically pleasing images and anchieve accurate emotion transfer, furthermore outperform the state-of-the-art methods. Our proposed EmoLat and Emospace Set have been publicly available to all scientific researchers for further deep research. In the future, we will focus on further exploring the potential of rich semantic relationships between objects in images for emotion space construction.

\section*{Acknowledgment}

This research is supported by Natural Science Foundation of Shanghai “Research on image sentiment analysis and expression based on human vision and cognitive psychology" under Grant No.22ZR1418400.

\bibliography{ref}
\bibliographystyle{IEEEtran}

\end{document}